\documentclass[letterpaper,onecolumn]{IEEEtran}

\title{Cumulative Regret Analysis of the Piyavskii--Shubert Algorithm and Its Variants for Global Optimization}
\author{
	Kaan Gokcesu, Hakan Gokcesu
}

\usepackage{latexsym,,amssymb,amsmath,graphicx,epsf,cite,float}
\usepackage{ifpdf}
\usepackage{epstopdf}

\usepackage{algorithm,algorithmic}
\usepackage{amsmath,amssymb,bm}
\usepackage{amsfonts,dsfont,color,subcaption}

\newcommand{\be}{\begin{equation}}
	\newcommand{\ee}{\end{equation}}
\newcommand{\bea}{\begin{eqnarray}}
	\newcommand{\eea}{\end{eqnarray}}

\newcommand{\MB}{\left[\begin{array}}
	\newcommand{\ME}{\end{array}\right]}

\newcommand{\ei}{\end{itemize}}
\newcommand{\bi}{\begin{itemize}}

\newcommand{\abs}[1]{|#1|}
\newcommand{\norm}[1]{\lVert#1\rVert}

\usepackage{amsmath,amsthm}
\newcommand\numberthis{\addtocounter{equation}{1}\tag{\theequation}}

\DeclareMathOperator*{\argmax}{arg\,max}

\newcommand{\E}{\mathbb{E}}

\usepackage[capitalize]{cleveref}

\newtheorem{result}{Result}
\newtheorem{theorem}{Theorem}

\newtheorem{lemma}[]{Lemma}

\newtheorem{proposition}[]{Proposition}

\newtheorem{corollary}[]{Corollary}

\newtheorem{remark}[]{Remark}

\newtheorem{definition}[]{Definition}

\usepackage[inline]{enumitem}
\usepackage{comment}

\begin{document}

	\maketitle

	\begin{abstract}
		We study the problem of global optimization, where we analyze the performance of the Piyavskii--Shubert algorithm and its variants. For any given time duration $T$, instead of the extensively studied simple regret (which is the difference of the losses between the best estimate up to $T$ and the global minimum), we study the cumulative regret up to time $T$. For $L$-Lipschitz continuous functions, we show that the cumulative regret is $O(L\log T)$. For $H$-Lipschitz smooth functions, we show that the cumulative regret is $O(H)$. We analytically extend our results for functions with Holder continuous derivatives, which cover both the Lipschitz continuous and the Lipschitz smooth functions, individually. We further show that a simpler variant of the Piyavskii-Shubert algorithm performs just as well as the traditional variants for the Lipschitz continuous or the Lipschitz smooth functions. We further extend our results to broader classes of functions, and show that, our algorithm efficiently determines its queries; and achieves nearly minimax optimal (up to log factors) cumulative regret, for general convex or even concave regularity conditions on the extrema of the objective (which encompasses many preceding regularities). We consider further extensions by investigating the performance of the Piyavskii-Shubert variants in the scenarios with unknown regularity, noisy evaluation and multivariate domain. 
	\end{abstract}
	
	\section{Introduction}
	In many applications such as hyper-parameter tuning for learning algorithms and complex system design, the goal is to optimize an unknown function with as few evaluations as possible and use that optimal point in the design \cite{poor_book, cesa_book}. In these types of problems, evaluating the performance of a set of parameters often requires numerical simulations (or cross-validations) with significant computational cost; and the constraints of the application often require a sequential exploration of the solution space with small number of evaluations. Moreover, the function to be optimized may not necessarily have nice properties such as linearity or convexity. 
	This generic problem of sequentially optimizing the output of an unknown and potentially non-convex function is often referred to as the following terminologies: global optimization \cite{pinter1991global}, black-box optimization \cite{jones1998efficient}, derivative-free optimization \cite{rios2013derivative}. 
	
	In the problem of global optimization, we have a function $f(\cdot)$ that we want to optimize. Let this function $f(\cdot)$ be univariate such that $f(\cdot)\colon \Theta\rightarrow\Omega,$ where $\Theta,\Omega \subset \Re$ and they are compact intervals. The goal of the global optimization is to find the optimizer of this function $f(\cdot)$ with as little evaluations of the function as possible. Every time a new point $x$ is queried, only its value $f(x)$, i.e., the result of the query, is revealed. Thus, for any	query $x\in\Theta,$ we observe $f(x)\in\Omega.$ Without loss of generality, we assume $\Theta=[0,1]$, since optimization on any compact interval $\Theta$ can be reduced to the problem of optimization on $[0,1]$ after translating and scaling of the input $x$. 
	
	\section{Related works}
	Global optimization has received considerable attention over the past decades. Many algorithms have been proposed in a myriad of fields such as convex optimization \cite{nesterov2003introductory,boyd2004convex,bubeck2015convex}, non-convex optimization \cite{hansen1992global,hansen1991number,jones1993lipschitzian, jain2017non, basso1982iterative, shang2019general}, stochastic optimization (or approximation) \cite{spall2005introduction,shalev2012online}, Bayesian optimization \cite{brochu2010tutorial}, bandit optimization over metric spaces \cite{munos2014bandits}. Additionally, due to numerous applications, it has been heavily utilized in decision theory \cite{tnnls4}, control theory \cite{tnnls3}, game theory \cite{tnnls1}, distribution estimation \cite{gokcesu2018density,willems,gokcesu2017online,coding2}, anomaly detection \cite{gokcesu2019outlier,gokcesu2016nested}, signal processing \cite{ozkan}, prediction \cite{singer,gokcesu2016prediction}, multi-armed bandits \cite{cesa-bianchi,gokcesu2018bandit,neyshabouri2018asymptotically}. 
	
	
	There exists various heuristics in literature that deal with this problem such as model-based methods, genetic algorithms and Bayesian optimization. However, the most popular approach is the regularity-based methods since, in many applications, the system has some inherent regularity in its objective, i.e., $f(\cdot)$ satisfies some regularity condition. 
	While there are works with different regularity conditions (e.g., \cite{bartlett2019simple} and \cite{grill2015black} use a notion of smoothness related to hierarchical partitioning), the most common is the Lipschitz continuity, where the function $f(\cdot)$ is Lipschitz continuous with some $L$.
	
	Lipschitz regularity was first studied in Piyavskii's work \cite{piyavskii1972algorithm}, which proposes a sequential deterministic method to solve the global optimization problem. The algorithm works by the iterative construction of a function $F(\cdot)$ that lower bounds the function $f(\cdot)$ and the evaluation of $f(\cdot)$ at a point where $F(\cdot)$ reaches its minimum. In the same year, Shubert has independently published the same algorithm \cite{shubert1972sequential}. Hence, this algorithm has been dubbed the Piyavskii--Shubert algorithm. 
	Basso \cite{basso1982iterative},
	Schoen \cite{schoen1982sequential}, Shen and Zhu \cite{shen1987interval}, Horst and Tuy \cite{horst1987convergence} propose other formulations of the Piyavskii’s algorithm. Sergeyev \cite{sergeyev1998global} use an alternative smooth auxiliary function as a lower bounding function.
	Hansen and Jaumard \cite{hansen1995lipschitz} summarize and discuss the algorithms proposed in the literature and present them in a high-level programming language. 
	Brent \cite{brent2013algorithms} studies another aspect of the Piyavskii’s algorithm, where the function is defined on a compact interval with a bounded second derivative (i.e., Lipschitz smoothness). Jacobsen and Torabi \cite{jacobsen1978global} assume that the function is differentiable and a sum of convex and concave functions. Mayne and Polak \cite{mayne1984outer} propose an outer approximation when the objective is non-differentiable. Mladineo \cite{mladineo1986algorithm} investigates the application on multimodal objectives. 
	Breiman and Cutler \cite{breiman1993deterministic} propose an approach, which use the Taylor expansion of $f(\cdot)$ to build a bounding function $F(\cdot)$. Baritompa and Cutler \cite{baritompa1994accelerations} propose an acceleration of the Breiman and Cutler’s method. Ellaia et al. \cite{ellaia2012modified} suggest a modified sequential Piyavskii’s algorithm.
	Horst and Tuy \cite{horst2013global} provides a general discussion on deterministic algorithms. 

	The performance analysis of optimization algorithms are generally done by the convergence to the optimizer. However, the convergence in the input domain becomes challenging when the objective function does not have nice properties like convexity. Hence, in global optimization, the convergence is studied with regards to the functional evaluation of the optimizer (i.e., the optimal value $\min_{x\in[0,1]}f(x)$).
	At each time $t$, the learner selects a point $x_t$ to be queried (evaluated). Generally, after each evaluation, the learner selects a point $x^*_t$, which may differ from the last queried point $x_t$. The accuracy of the approximation provided by the point $x^*_t$ (returned after the $t^{th}$ evaluation of the function) is measured with its closeness to the optimal function evaluation, i.e., the simple regret at evaluation $T$ is given by the difference of the evaluation of the estimation $x^*_T$ and the optimal point (global minimizer) $x^*$.
	Although the Piyavskii--Shubert algorithm was extensively studied in the literature, its regret analysis was limited. The study of the number of iterations by the Piyavskii--Shubert algorithm was initiated by Danilin \cite{danilin1971estimation}. For its simple regret analysis (which is the difference between the best evaluation so far and the optimal evaluation), a crude regret bound of the form $\tilde{r}_T = O(T^{-1})$ is derived by Mladineo \cite{mladineo1986algorithm} when the function $f(\cdot)$ is Lipschitz continuous. The authors further show that the Piyavskii–Shubert algorithm is minimax optimal and superior to uniform grid search. 
	For univariate functions, a bound on the evaluation complexity was derived by Hansen et al. \cite{hansen1991number} for a variant of the Piyavskii–Shubert algorithm that stops automatically after returning an $\epsilon$-optimal of $f(\cdot)$. They proved that the number of iterations required to reach precision $\epsilon$ is at most proportional to $\int_0^1(f(x)-f(x^*)+\epsilon)^{-1}dx$. For these results, the authors explicitly study the lower bounding functions and improve upon the results of \cite{danilin1971estimation}. The work of Ellaia et al. \cite{ellaia2012modified} further improves upon the results of \cite{danilin1971estimation,hansen1991number}.
	A work by Malherbe and Vayatis \cite{malherbe2017global} studied a variant of the Piyavskii–Shubert algorithm called LIPO. Rather than optimizing a single proxy function,
	LIPO queries the next point randomly in a set of potential optimizers. Their regret bounds depend on stronger assumptions. The work by Bouttier et al. \cite{bouttier2020regret} studies the simple regret of Piyavskii--Shubert under noisy evaluations. Furthermore, \cite{bouttier2020regret} and Bachoch et al. \cite{bachoc2021instance} show that the Piyavskii--Shubert algorithm has instance-optimal simple regret guarantees.
	
	\section{Main results and contributions}
	
	As in the learning literature, we consider $f(\cdot)$ as a loss function that we want to minimize by choosing estimations $x_t$ from $\Theta$ at each time $t$. The goal is to design an iterative algorithm that creates new estimations using the previously acquired information, i.e., ${x_t=\Gamma_t(f(x_1),\ldots,f(x_{t-1}), x_1,\ldots,x_{t-1})},$
	where $\Gamma(\cdot)$ is some mapping, and the initial selection $x_1$ is predetermined. 
	It is not straightforward to optimize any arbitrary $f(\cdot)$. To this end, we define a regularity measure. Instead of the restrictive Lipschitz continuity; we define a weaker, more general regularity condition.
	\begin{definition}\label{def:condition}
		Let the function $f(\cdot)$ that we want to optimize satisfy the following condition: $|f(x)-f(x_E)|\leq d\left(\abs{x-x_E}\right),$
		for any local extremum (minimum or maximum) $x_E$ of $f(\cdot)$, where $d(\cdot)\colon [0,1]\rightarrow \Re$ is a known monotone nondecreasing function that satisfies $d(0)=0$.
	\end{definition}
	
	A preliminary contribution of our work is the regularity in \cref{def:condition}.
	There are two key points: \begin{itemize*}
		\item Instead of the traditional Lipschitz continuity with the absolute function $|\cdot|$, we analyze a general regularity $d(\cdot)$, which covers other regularities such as Lipschitz smoothness and Holder continuity.
		\item The regularity in \cref{def:condition} is only on the extrema of $f(\cdot)$ instead of the whole function, which is especially meaningful when $d(\cdot)$ is convex. If a function $f(\cdot)$ satisfies a convex regularity at every point on its domain, it may only be a constant trivial function. On the other hand, with the extrema regularity of \cref{def:condition}, we can consider a multitude of functions, e.g., periodic functions.
	\end{itemize*} 
	
	Since the objective is treated as a loss function, the goal of an algorithm is to minimize the evaluation $f(x_t)$ observed at time $t$. Since the functional evaluations can be arbitrarily high, we use the notion of regret to define our algorithmic performance. Let the global minimum of $f(\cdot)$ be $f_*\triangleq \min_{x\in\Theta} f(x).$ Hence, the goal is to make the difference $f(x_t)-f_*$ as small as possible. Instead of the weaker simple regret, for the first time in the literature, we study the stronger cumulative regret for global optimization. Thus, instead of the simple regret $r_T\triangleq\min_{t\in\{1,\ldots,T\}}f(x_t)-f_*,$ we analyze the cumulative regret up to time $T$, which is $R_T\triangleq\sum_{t=1}^{T}f(x_t)-\sum_{t=1}^{T}f_*\label{eq:regret}.$ There exist two key points: \begin{itemize*}
		\item Firstly, cumulative regret is more meaningful when there are operational costs driven by evaluation.
		\item Secondly, cumulative regret may be more indicative of algorithmic performance, since the trivial grid search has an abysmal linear cumulative regret in spite of its minimax optimal simple regret.
	\end{itemize*}
	
	For the first time in the literature, we derive cumulative regret bounds for the famous traditional Piyavskii--Shubert algorithm for various polynomial regularities. These bounds have alluded the global optimization field for many years. We were able to achieve them by showing recursive relations for functions that upper-bound the regrets. 
	\begin{result}[Regret of Traditional Piyavskii--Shubert Algorithm]
		When the regularity function is $d(\cdot)=K|\cdot|^p$ for $K>0,p\geq 1$; the Piyavskii--Shubert variants have the cumulative regret $R_T\leq 2K\min\left(\log(4T),({1-\gamma})^{-1}\right),$ where $\gamma\triangleq\left({2^{p}-1}\right)^{-1}\leq 1$. 
		Specifically; 
		when $p=1$, we have $R_T\leq O(K\log(T));$
		when $p=2$, we have $R_T\leq 2K;$
		for all $p\geq 2$, we have $R_T\leq 3K;$
		for any $p>1$, we have $R_T\leq O(K)$, i.e., independent from $T$.
	\end{result}
	
	We show that a simpler Piyavskii--Shubert algorithm performs just as well as the traditional variants. It is motivated from the fact that an inherent issue with the traditional variants is the lack of control on the iterative optimization regions. Because of its simpler structure, we are able to provide cumulative regret guarantees for a larger variety of regularities, including general convex and concave regularities.
	\begin{result}[Regret of Simpler Piyavskii--Shubert Algorithm]
		When the regularity is $d(\cdot)=K|\cdot|^p$ for $K>0,p> 0$; the simpler variants have the cumulative regret $R_T \leq K+2K\frac{2^{(1-p)\log(2T)}-1}{2^{1-p}-1}.$ 
		Specifically; when $p=1$: we have $R_T\leq O(K\log T)$; for all $p\geq2$, we have $R_T\leq 5K$; for any $p>1$: we have $R_T\leq O(K)$; when $0<p<1$: we have $R_T\leq O(KT^{1-p})$.
	\end{result}
	
	We show that all of these regret bounds are minimax optimal up to logarithmic terms. Moreover, we also investigate various extensions in our analyses. 
	When the regularity is unknown, we show that either the regret bound is a piecewise linear function of the input regularity coefficient or the logarithmic regret bound is a piecewise linear function of the logarithm of the input regularity coefficient depending on whether the unknown regularity satisfies some properties such as convexity or concavity. We also show how to deal with noisy evaluations and their effect on the regret bounds. We also show that the univariate optimization can be readily applied to the multivariate case with minimax optimal regret guarantees when the regularity is on the whole function.

	\section{Global optimization algorithmic framework}\label{sec:algorithm}
	In this section\footnote{We use $\log(\cdot)$ to denote the base-$2$ logarithm.} , we provide the algorithmic framework for various Piyavskii--Shubert variants, which can efficiently optimize the objective function $f(\cdot)$ with low regret bounds.
	Piyavskii–Shubert variants generally work by creating a function $F(\cdot)$, which globally lower bounds the objective function $f(\cdot)$. At each time $t$, the lower bounding function $F(\cdot)$ is updated (or recreated), hence, we have a time varying function $F_t(\cdot)$. Each $F_t(\cdot)$ is created by using the information of the past evaluations $\{f(x_\tau)\}_{\tau=1}^{t-1}$ and the query points $\{x_\tau\}_{\tau=1}^{t-1}$. Each estimation $x_t$ are selected from the global minima of the lower bounding function $F_t(\cdot)$.
	Based on this methodology, we provide a general global optimization algorithmic framework in \cref{alg}.
	\begin{algorithm}[!t]
		\caption{Global Optimization Algorithmic Framework}\label{alg}
		\normalsize{\begin{algorithmic}[1]
				\STATE Take the function to be optimized $f(\cdot):[0,1]\rightarrow\Re$ and the regularity parameters as input. 
				\STATE Initialize the candidate set $\mathcal{X}=\emptyset$ and the evaluated set $\mathcal{Y}=\emptyset$.
				\STATE Evaluate $f(x)$ at $x=0$. Evaluate $f(x)$ at $x=1$. \\Add to $\mathcal{Y}$ the evaluation pairs $(0,f(0))$ and $(1,f(1))$.
				\STATE If possible, determine a candidate $x\in(0,1)$ and its score $s$ with the evaluations $\{f(0),f(1)\}$. \\If $s$ is lower than the minimum evaluation in $\mathcal{Y}$, add $(x,s)$ to $\mathcal{X}$.  \\The score $s$ determines which candidate to evaluate.\label{item:candidate}
				\WHILE{$|\mathcal{X}|\neq \emptyset$}
				\STATE Evaluate the point $x_m$ with the minimum score $s_m$ in $\mathcal{X}$. \\Remove $(x_m,s_m)$ from $\mathcal{X}$. \\Add $(x_m,f(x_m))$ to $\mathcal{Y}$. \\Let $x_m$ be the only point between the previously evaluated points $x_l$ and $x_r$ (i.e., $x_m\in(x_l,x_r)$).
				\STATE If possible, determine a candidate $x_{lm}\in(x_l,x_m)$ and its score $s_{lm}$ with the respective evaluations $\{f(x_l),f(x_m)\}$.\label{step:xlm} \\
				If $s_{lm}$ is lower than the minimum evaluation in $\mathcal{Y}$, add $(x_{lm},s_{lm})$ to $\mathcal{X}$.
				\STATE If possible, determine a candidate $x_{mr}\in(x_m,x_r)$ and its score $s_{mr}$ with the respective evaluations $\{f(x_m),f(x_r)\}$.\label{step:xmr} \\
				If $s_{mr}$ is lower than the minimum evaluation in $\mathcal{Y}$, add $(x_{mr},s_{mr})$ to $\mathcal{X}$.
				\ENDWHILE
				\WHILE{\TRUE}
				\STATE Evaluate the point $x$ with minimum evaluation $f(x)$ in $\mathcal{Y}$.
				\ENDWHILE	\end{algorithmic}}
	\end{algorithm}
	There exist two key points in this algorithm: 
	\begin{itemize*}
		\item Firstly, instead of creating the whole lower bounding function $F_t(\cdot)$, our framework only creates the relevant minima and their function evaluations in $F_t(\cdot)$, i.e., the next query $x_t$ and its score $s_t$. 
		\item Secondly, the algorithm updates the lower bounding function by only using the newly evaluated points and their evaluations. The lower bounding functions are created in a piecewise manner using the respective boundary evaluations. 
	\end{itemize*}
	We observe that the creation of the proxy function itself is inconsequential and the important aspect is determining its extrema. In another perspective, we postulate that it comes down to determining some candidate points and lower bounding scores. 
	
	\begin{remark}
		\cref{alg} is equal to the traditional Piyavskii--Shubert algorithm when the query point is chosen as the intersection of lines with reverse slopes from its respective boundary evaluations. 
	\end{remark}
	\begin{remark}
		During the run of the algorithm, let the number of evaluations so far be $T$. After each evaluation, at most two potential query points are iteratively created; and the queries not evaluated remain unchanged. The time complexity per evaluation is at most $O(\log T)$. Between any adjacent queried points, there will be at most one potential query. Hence, the number of potential queries grows at most linearly with the number of queries. Thus, memory complexity is at most $O(T)$.
	\end{remark}
	
	\begin{remark}
		The algorithm naturally terminates when there are no queries left in the query list. Additional stopping criteria can also be considered, such as stopping the algorithm after a fixed amount of trials or when a sufficient closeness to the optimizer value is reached, which measured as the difference of the minimum evaluation so far and currently minimum score $\mathcal{X}$. 
	\end{remark}
	
	\begin{remark}
		We can further increase computational efficiency by eliminating the potential queries with scores that are higher than the minimum function evaluation queried so far. 
	\end{remark}
	
	\section{Regret analysis of the traditional Piyavskii--Shubert variants}\label{sec:traditional}
	In this section, we investigate the cumulative regret of the traditional Piyavskii--Shubert algorithm and its variants. The candidate point $x$ (in Step \ref{item:candidate}, \ref{step:xlm}, \ref{step:xmr} of \cref{alg}) is selected as the point with the lowest possible functional value between the nearest boundaries $x_l$ and $x_r$. Its score $s$  is set as its lowest possible evaluation. Next, we analyze different classes of commonly utilized functions.
	\subsection{Optimizing Lipschitz continuous functions}\label{sec:LcontPS}
	A function $f(\cdot)$ is Lipschitz continuous with $L$ if $\abs{f(x)-f(y)}\leq L\abs{x-y},\forall x,y \in \Theta.$
	We can see that for any extremum $x_E$ of $f(\cdot)$, we have $|f(x)-f(x_E)|\leq L|x-x_E|.$ Here, the candidate point $x$ is the intersection of the lines that pass through points $(x_l,f_l)$, $(x_r,f_r)$ with slopes $-L$, $L$, respectively. Using the following $(x,s)$ in \cref{alg} is the traditional Piyavskii--Shubert algorithm.
	\begin{lemma}\label{thm:candidate}
		For an $L$-Lipschitz continuous function $f(\cdot)$, given the boundaries $x_l$, $x_r$ and evaluations $f_l\triangleq f(x_l)$, $f_r\triangleq f(x_r)$; the candidate $x\in(x_l,x_r)$ with the lowest possible value and its score are 
		\begin{align}
			&\textstyle x=0.5\left(x_l+x_r+{({f_l-f_r})}/{L}\right),\\
			&\textstyle s=0.5\left(f_l+f_r-L(x_r-x_l) \right).
		\end{align}
	\end{lemma}
	\begin{proof}
		Since the point $x$ is the intersection of the lines that pass through points $(x_l,f_l)$, $(x_r,f_r)$ with slopes $-L$, $L$ respectively. We have the following equality $$f_l+(-L)(x-x_l)+L(x_r-x)=f_r,$$
		which results in the candidate point $$x=\frac{1}{2}\left(x_r+x_l+\frac{{f_l-f_r}}{L}\right).$$
		Moreover, we have $$s=f_l+(-L)(x-x_l).$$ Putting in the expression for $x$, we get $$s=\frac{1}{2}\left(f_r+f_l-L(x_r-x_l) \right)$$ and conclude the proof.
	\end{proof}
	
	We next study the cumulative regret of \cref{alg} when the candidate points and scores are selected with respect to \cref{thm:candidate}. We start by investigating the regret of a single query.
	
	\begin{lemma}\label{thm:sampleRegretPSL}
		For an $L$-Lipschitz continuous function $f(\cdot)$, let $x_m$ be the next query between the boundaries $x_l$, $x_r$ and their evaluations $f_l\triangleq f(x_l)$, $f_r\triangleq f(x_r)$. the regret incurred by $x_m$ is
		\begin{align}
			\textstyle f(x_m)-\min_{x\in[0,1]}f(x)\leq 2L\min(x_m-x_l, x_r-x_m).
		\end{align} 
	\end{lemma}
	\begin{proof}
		Since the score $s_m$ of the candidate $x_m$ is a minimizer for the lower bounding function between $[x_l,x_r]$, we have $$s_m\triangleq\frac{1}{2}\bigg(f_r+f_l-L(x_r-x_l)\bigg)\leq \min_{x\in[x_l,x_r]}f(x).$$
		Because of the way the algorithm works, an evaluation in each interval is either lower bounded by a candidate score or a past evaluation. Since we evaluate the candidate point with the lowest score that is also lower than the past evaluations, we have
		\begin{align}
			s_m\triangleq\frac{1}{2}\bigg(f_r+f_l-L(x_r-x_l)\bigg)\leq \min_{x\in[0,1]}f(x).
		\end{align}
		Moreover, because of Lipschitz continuity, we have
		\begin{align}
			f(x_m)\leq& f_l+f_l-s_m,\\
			\leq& f_r+f_r-s_m,\\
			\leq& 2\min(f_l,f_r)-s_m.
		\end{align}
		Thus,
		\begin{align}
			f(x_m)-\min_{x\in[0,1]}f(x)&\leq 2\min(f_l,f_r)-2s_m,\\
			&\leq L(x_r-x_l)-\abs{f_r-f_l}.	
		\end{align}
		Then, using \cref{thm:candidate}, we substitute $\abs{f_r-f_l}$ with the equivalent expression $L\abs{(x_r-x_m)-(x_m-x_l)}$, i.e., 
		\begin{align}
			f(x_m)-\min_{x\in[0,1]}f(x)&\leq L(x_r-x_l)-L\abs{(x_r-x_m)-(x_m-x_l)}.	
		\end{align}
		If $(x_r-x_m)-(x_m-x_l)\geq0,$ we have
		\begin{align}
			f(x_m)-\min_{x\in[0,1]}f(x)\leq& L(x_r-x_l)-L({(x_r-x_m)-(x_m-x_l)}),
			\\\leq&	2L(x_m-x_l).
		\end{align}
		On the other hand, if $(x_r-x_m)-(x_m-x_l)<0,$ we have
		\begin{align}
			f(x_m)-\min_{x\in[0,1]}f(x)\leq& L(x_r-x_l)+L({(x_r-x_m)-(x_m-x_l)}),
			\\\leq&	2L(x_r-x_m).
		\end{align}
		Hence,
		\begin{align}
			f(x_m)-\min_{x\in[0,1]}f(x)\leq&	2L\min(x_m-x_l,x_r-x_m),
		\end{align}
		which concludes the proof.
	\end{proof}
	
	The result in \cref{thm:sampleRegretPSL} bounds the single query regret with respect to the minimum distance to the previously evaluated boundaries.
	Let us observe that the lower bounding algorithm in \cref{alg} is deterministic. Hence, given $L$, function $f(\cdot)$ and interval $[0,1]$ (where $0$ and $1$ are initially evaluated), it will always evaluate the same points. Let these be $x_1^T$, i.e., $x_1^T=PS_{LiCo}([0,1],f(\cdot),L,T),$ for some function $PS_{LiCo}(\cdot)$, which represents the application of \cref{alg} to $L$-Lipschitz continuous functions with the selections in \cref{thm:candidate}. Let $S_L([0,1], x_1^T)=2L\sum_{t=1}^T\min_{-1\leq\tau\leq t-1}|x_t-x_\tau|,$ be the cumulative bound resulting from \cref{thm:sampleRegretPSL}
	where $x_{0}=0$, $x_{-1}=1$ are initially evaluated boundary points of $[0,1]$. Let the maximum bound be 
	\begin{align}
		&\textstyle R_L([0,1],T)\\\numberthis
		&=\underset{\substack{0\leq l\leq L\\f(\cdot):\text{$l$-L. cont.}}}{\max}S_L([0,1],PS_{LiCo}([0,1],f(\cdot),l,T)).\label{eq:regC}
	\end{align}
	\begin{lemma}\label{thm:rec}
		We have the following recursive relation for the maximum bound defined in \eqref{eq:regC}
		\begin{align}
			R_L([0,1],T)\leq& 2L\min(x,1-x)+xR_L([0,1],T_1)\\
			&+(1-x)R_L([0,1],T_2)
		\end{align}
		for some $x\in[0,1]$ and $T_1+T_2=T-1$.
	\end{lemma}
	\begin{proof}
		Let $$(f^*(\cdot),l^*)\in\argmax_{(f(\cdot),l)}S_L([0,1],PS_{LiCo}([0,1],f(\cdot),l,T))$$ be some minimizing function and parameter pair. Let $${x}_1^T=PS_{LiCo}([0,1],f^*(\cdot),l^*,T)$$ be the lower bounding algorithms point selections given $f^*(\cdot)$ and $l^*$.
		Let $$y_1^{T_1}=\{y:y\in x_1^T,y<x_1\}$$ and $$z_1^{T_2}=\{z:z\in x_1^T,z>x_1\},$$ whose elements have the same ordering as in $x_1^T$, individually. Then, we have 
		\begin{align}
			R_L([0,1],T)=2L\min(x_1,1-x_1)+S_L([0,x_1], y_1^{T_1})+S_L([x_1,1], z_1^{T_2}).\label{eq:RS}
		\end{align}
		From the deterministic nature and working structure of the lower bounding algorithms where the candidate selections are done on the relative disjoint regions separately, we have
		\begin{align}
			S_L([0,x_1], y_1^{T_1})=&S_L([0,x_1],PS_{LiCo}([0,x_1],f^*(\cdot),l^*,T_1))\label{eq:deter1}
			\\=&S_L([0,x_1],x_1PS_{LiCo}([0,1],f^*(x_1(\cdot)),x_1l^*,T_1))\label{eq:deter2}
			\\=&x_1S_L([0,1],PS_{LiCo}([0,1],f^*(x_1(\cdot)),x_1l^*,T_1))\label{eq:defS}
			\\=&x_1S_L([0,1],PS_{LiCo}([0,1],f'(\cdot),l',T_1))\label{eq:rep}
			\\\leq& x_1R_L([0,1],T_1)\label{eq:defR},
		\end{align}
		where \eqref{eq:defS} comes from the linearity with respect to the sample points due to the definition of the function $S_L(\cdot,\cdot)$; in \eqref{eq:rep} $f^*(x_1(\cdot))$ is replaced with $f'(\cdot)$ and $x_1l^*$ is replaced with $l'$; and \eqref{eq:defR} comes from the definition of the function $R_L(\cdot,\cdot)$.
		
		Similar derivations hold for $$S_L([x_1,1], z_1^{T_2})\leq (1-x_1)R_L([0,1],L,T_2).$$ When both of them are substituted in \eqref{eq:RS}, we reach the intended result.
	\end{proof}
	
	Using the recursion in \cref{thm:rec}, we reach the following regret bounds.
	\begin{theorem}\label{thm:regretLPS}
		For an $L$-Lipschitz continuous function $f(\cdot)$, the traditional Piyavskii--Shubert variant with its application in \cref{thm:candidate} has the cumulative regret
		$R_T\leq 2L\log(4T).
		$
	\end{theorem}
	\begin{proof}
		We run the algorithm for $T$ evaluations. Assume at first $T\geq 3$. After evaluating the boundary points $x=0$ and $x=1$, there exist $T-2$ more evaluations. We have
		\begin{align}
			R_T\leq R_L([0,1],T-2)+L.
		\end{align}
		Let the next evaluated point after $\{0,1\}$ be $x$. From \cref{thm:rec}, $\exists T_1, T_2$, we have
		\begin{align}
			R_L([0,1],T-2)\leq 2L\min(x,1-x)&+xR_L([0,1],T_1)+(1-x)R_L([0,1],T_2),
		\end{align}
		for some $x\in[0,1]$ and $T_1+T_2=T-3$.
		If $x\leq1-x$, we can set $T_1\leq T_2$ to upper bound the regret since $R_L(\cdot,\cdot)$ is nondecreasing with respect to its second argument. Hence, $T_1\leq \lfloor {(T-2)/2}\rfloor$, $T_2\leq T-2$ (we can do the converse if $1-x<x$). Thus, we have the following, where $\bar{x}=\min(x,1-x)$,
		\begin{align}
			R_L([0,1],T-2)\leq& 2L\bar{x}+\bar{x}R_L([0,1],\lfloor(T-2)/2\rfloor)+(1-\bar{x})R_L([0,1],T-2),\\
			\bar{x}R_L([0,1],T-2)\leq&2L\bar{x}+\bar{x}R_L([0,1],\lfloor(T-2)/2\rfloor),\\
			R_L([0,1],T-2)\leq&2L+R_L([0,1],\lfloor(T-2)/2\rfloor),
		\end{align}
		since $\bar{x}>0$ (if not, candidate point is unavailable, i.e., we found the global minimum). The recursive rule gives
		\begin{align}
			R_L([0,1],T-2)\leq& 2L\log(2T)+R_L([0,1],1),
			\\\leq& 2L\log(2T)+L,
		\end{align}
		where $\log(\cdot)$ is in base $2$.
		Hence, with the initial evaluations $x=0$ and $x=1$, we have
		\begin{align}
			R_T&\leq 2L\log(2T)+2L\leq 2L\log(4T). 
		\end{align}
	\end{proof}
	
	
	\subsection{Optimizing Lipschitz smooth functions}\label{sec:LsmoothPS}
	A function $f(\cdot)$ is Lipschitz smooth with $H$ if $\abs{f'(x)-f'(y)}\leq H\abs{x-y}.$ We can see that for any extremum $x_E$ of $f(\cdot)$, we have $|f(x)-f(x_E)|\leq 0.5H|x-x_E|^2,$
	since the derivative is zero at any $x_E$.
	Here, the candidate point $x$ (if exists) is the minimum point of a quadratic function that passes through points $(x_l,f_l)$, $(x_r,f_r)$ with the second derivative $H$. $(x,s)$ is given as in the following.
	\begin{lemma}\label{thm:candidateH}
		For a $H$-Lipschitz smooth function $f(\cdot)$, given the boundaries $x_l$, $x_r$ and evaluations $f_l\triangleq f(x_l)$ and $f_r\triangleq f(x_r)$, the candidate $x$ with the lowest possible value and its score are
		\begin{align}
			&\textstyle x=\frac{x_l+x_r}{2}+\frac{{f_l-f_r}}{H(x_r-x_l)},\\
			&\textstyle s=f_l-\frac{1}{2}H(x-x_l)^2=f_r-\frac{1}{2}H(x_r-x)^2.
		\end{align}
		If $x\notin(x_l,x_r)$, there exists no candidate in $[x_l,x_r]$, i.e., $f(x)\geq \min(f(x_l),f(x_r)), \forall x\in[x_l,x_r]$.
	\end{lemma}
	\begin{proof}
		Since the point $x$ is the minimum point of the quadratic function that passes through points $(x_l,f_l)$, $(x_r,f_r)$ with the second derivative $H$. We have the following equality
		\begin{align}
			f_l-\frac{1}{2}H(x-x_l)^2+\frac{1}{2}H(x_r-x)^2=f_r, &&x\in[x_l,x_r].
		\end{align} 
		This gives
		\begin{align}
			f_l-\frac{1}{2}H(x_l^2-2xx_l)+\frac{1}{2}H(x_r^2-2xx_r)=f_r.
		\end{align}
		Hence,
		\begin{align}
			Hx(x_r-x_l)=&\frac{1}{2}H(x_l+x_r)(x_r-x_l)+f_l-f_r,\\
			x=&\frac{x_l+x_r}{2}+\frac{f_l-f_r}{H(x_r-x_l)},
		\end{align}	
		which concludes the proof.
	\end{proof}
	We next study the cumulative regret of \cref{alg} when the candidate points and scores are selected with respect to \cref{thm:candidateH}. We start by investigating the regret of a single query.
	
	\begin{lemma}\label{thm:sampleRegretH}
		For an $H$-Lipschitz smooth function $f(\cdot)$, let $x_m$ be the next query, which is between the boundary points $x_l$, $x_r$ and their corresponding values $f_l\triangleq f(x_l)$, $f_r\triangleq f(x_r)$. The regret incurred by the evaluation of $x_m$ is
		\begin{align}
			\textstyle f(x_m)-\min_{x\in[0,1]}f(x)\leq H(x_r-x_m)(x_m-x_l).
		\end{align} 
	\end{lemma}
	\begin{proof}
		From the derivation of the score $s_m$ of $x_m$, we know that $$s_m\triangleq \min(f_l,f_r)-\frac{1}{2}H\bar{x}^2\leq \min_{x\in[x_l,x_r]}f(x),$$
		where $\bar{x}=\min(x_m-x_l,x_r-x_m).$ Because of the way the algorithm works, an evaluation in each interval is either lower bounded by candidate scores or already evaluated points. Since we evaluate the candidate point with the lowest score that is also lower than the past evaluations, we have $$s_m\leq \min_{x\in[0,1]}f(x).$$
		Moreover, because of Lipschitz smoothness, we have
		\begin{align}
			f(x_m)\leq& f_l+\frac{1}{2}H(x_r-x_m)^2-\frac{1}{2}H(x_r-2x_m+x_l)^2,\\
			f(x_m)\leq& f_r+\frac{1}{2}H(x_l-x_m)^2-\frac{1}{2}H(x_r-2x_m+x_l)^2.
		\end{align}
		Thus,
		\begin{align}
			f(x_m)\leq& \min(f_l+\frac{1}{2}H(x_r-x_m)^2,f_r+\frac{1}{2}H(x_l-x_m)^2)-\frac{1}{2}H(x_r-2x_m+x_l)^2.
		\end{align}
		Using $\bar{x}=\min(x_m-x_l,x_r-x_m)$ and moving the secondary terms out of $\min(\cdot,\cdot)$ by upper-bounding, we have the following expression
		\begin{align}
			f(x_m)\leq& \min(f_l,f_r)+\frac{1}{2}H((x_r-x_l-\bar{x})^2-(x_r-x_l-2\bar{x})^2).
		\end{align}
		Finally, from $$s_m\triangleq\min(f_l,f_r)-\frac{1}{2}H\bar{x}^2\leq \min_{x\in[0,1]}f(x),$$
		we have
		\begin{align}
			\min(f_l,f_r)\leq \min_{x\in[0,1]}f(x)+\frac{1}{2}H\bar{x}^2,
		\end{align}
		which results in
		\begin{align}
			f(x_m)-\min_{x\in[0,1]}f(x)&\leq \frac{1}{2}H(\bar{x}^2+(x_r-x_l-\bar{x})^2-(x_r-x_l-2\bar{x})^2),\\
			&\leq H(\bar{x}(x_r-x_l-\bar{x})),\\
			&\leq H(x_r-x_m)(x_m-x_l),				
		\end{align}
		and concludes the proof.
	\end{proof}
	\cref{thm:sampleRegretH} bounds the single query regret with respect to the multiplication of distances to the boundaries. Similar to \cref{thm:rec}, we have a recursive inequality for an upper bound of the regret.
	\begin{lemma}\label{thm:recH}
		We have the following recursive relation for an upper bound on the regret
		\begin{align}
			\textstyle 	R_H([0,1],T)\leq& Hx(1-x)+x^2R_H([0,1],T_1)\\
			&+(1-x)^2R_H([0,1],T_2)
		\end{align}
		for some $x\in[0,1]$ and $T_1+T_2=T-1$.
	\end{lemma}
	\begin{proof}
		Let us observe that the lower bounding algorithm in \cref{alg} is deterministic. Hence, given $H$, function $f(\cdot)$ and interval $[0,1]$ (where $0$ and $1$ are already evaluated), it will always evaluate the same points. Let these be $x_1^T$, i.e., $$x_1^T=PS_{LiSm}([0,1],f(\cdot),H,T),$$ for some function $PS_{LiSm}(\cdot)$, which represents the application of \cref{alg} to Lipschitz smooth functions. Let $S_H([0,1], x_1^T)$ be the cumulative bound resulting from \cref{thm:sampleRegretH} $$S_H([0,1], x_1^T)=H\sum_{t=1}^T\left(\min_{x:x\in\{x_{-1}^{t-1}\},x<x_t}(x_t-x)\right)\left(\min_{x:x\in\{x_{-1}^{t-1}\}, x>x_t}(x-x_t)\right),$$
		where $x_{0}=0$, $x_{-1}=1$ are initially evaluated boundary points of $[0,1]$. Let the maximum bound be $$R_H([0,1],T)=\max_{0\leq h\leq H}\max_{f(\cdot):\text{$h$-Lipschitz smooth}}S_H([0,1],PS_{LiSm}([0,1],f(\cdot),h,T)).$$ 
		The rest of the proof is similar to the proof of \cref{thm:rec}.
		
		Let $$(f^*(\cdot),h^*)\in\argmax_{(f(\cdot),h)}S_H([0,1],PS_{LiSm}([0,1],f(\cdot),h,T))$$ be some minimizing function and parameter pair. Let $${x}_1^T=PS_{LiSm}([0,1],f^*(\cdot),h^*,T)$$ be the lower bounding algorithms point selections given $f^*(\cdot)$ and $h^*$.
		Let $$y_1^{T_1}=\{y:y\in x_1^T,y<x_1\}$$ and $$z_1^{T_2}=\{z:z\in x_1^T,z>x_1\},$$ whose elements have the same ordering as in $x_1^T$, individually. Then, we have 
		\begin{align}
			R_H([0,1],T)=Hx_1(1-x_1)+S_H([0,x_1], y_1^{T_1})+S_H([x_1,1], z_1^{T_2}).\label{eq:RSH}
		\end{align}
		From the deterministic nature and working structure of the lower bounding algorithms where the candidate selections are done on the relative disjoint regions separately, we have
		\begin{align}
			S_H([0,x_1], y_1^{T_1})=&S_H([0,x_1],PS_{LiSm}([0,x_1],f^*(\cdot),h^*,T_1))\label{eq:deter1H}
			\\=&S_H([0,x_1],x_1PS_{LiSm}([0,1],f^*(x_1(\cdot)),x_1^2h^*,T_1))\label{eq:deter2H}
			\\=&x_1^2S_H([0,1],PS_{LiSm}([0,1],f^*(x_1(\cdot)),x_1^2h^*,T_1))\label{eq:defSH}
			\\=&x_1^2S_H([0,1],PS_{LiSm}([0,1],f'(\cdot),h',T_1))\label{eq:repH}
			\\\leq& x_1^2R_H([0,1],T_1)\label{eq:defRH},
		\end{align}
		where \eqref{eq:defSH} comes from the quadratic dependency with respect to the sample points due to the definition of the function $S_H(\cdot,\cdot)$; in \eqref{eq:repH} $f^*(x_1(\cdot))$ is replaced with $f'(\cdot)$ and $x_1^2h^*$ is replaced with $h'$; and \eqref{eq:defRH} comes from the definition of the function $R_H(\cdot,\cdot)$.
		
		Similar derivations hold for $$S_H([x_1,1], z_1^{T_2})\leq (1-x_1)^2R_H([0,1],T_2).$$ When both of them are substituted in \eqref{eq:RSH}, we reach the intended result.
	\end{proof}
	
	Using the recursion in \cref{thm:recH}, we reach the following constant cumulative regret bound.
	\begin{theorem}\label{thm:regretHPS}
		For an $H$-Lipschitz smooth function $f(\cdot)$, the traditional Piyavskii--Shubert variant with its application in \cref{thm:candidateH} has the cumulative regret $R_T\leq H.$
	\end{theorem}
	\begin{proof}
		We run the algorithm for $T$ evaluations. After evaluating the boundary points $x=0$ and $x=1$, there exist $T-2$ more evaluations. We have
		\begin{align}
			R_T\leq R_H([0,1],T-2)+\frac{1}{2}H.
		\end{align}
		Let the next evaluated point after $\{0,1\}$ be $x$. We have from \cref{thm:recH}
		\begin{align}
			R_H([0,1],T-2)\leq& Hx(1-x)+x^2R_H([0,1],T_1)+(1-x)^2R_H([0,1],T_2).
		\end{align}
		We can set $T_1\leq T-2$ and $T_2\leq T-2$ since $T_1+T_2=T-3$, which gives
		\begin{align}
			R_H([0,1],T-2)\leq& Hx(1-x)+x^2R_H([0,1],T-2)+(1-x)^2R_H([0,1],T-2).
		\end{align}
		After rearranging, we get
		\begin{align}
			(2x-2x^2)R_H([0,1],T-2)\leq& Hx(1-x).
		\end{align} 
		Since $0< x< 1$ (otherwise we have a global minimum) and $x(1-x)> 0$, we have
		\begin{align}
			R_H([0,1],T-2)&\leq \frac{1}{2}H.
		\end{align}
		Adding the initial evaluations $x=0$ and $x=1$, we have $R_T\leq H,$ which concludes the proof.
	\end{proof}
	
	\subsection{Optimizing Holder smooth functions}\label{sec:LanalyticPS}
	Here, we aim to optimize a $K$-Holder smooth function $f(\cdot)$ with Holder continuous derivative, i.e., $\abs{f'(x)-f'(y)}\leq Kp\abs{x-y}^{p-1},$
	where $0\leq p-1\leq 1$. 
	When $p=1$, this includes the Lipschitz continuous functions in \cref{sec:LcontPS} with $K=L$. 
	When $p=2$, this class is the Lipschitz smooth functions in \cref{sec:LsmoothPS} with $2K=H$. 
	We have $|f(x)-f(x_E)|\leq K|x-x_E|^p,$
	where $x_E$ is any local extremum of $f(\cdot)$, $K>0$ and $1\leq p\leq 2$. 
	The candidate $x$ (if exists) is the minimum point of a convex $p^{th}$-order polynomial that passes through points $(x_l,f_l)$, $(x_r,f_r)$ with the coefficient $K$. 
	\begin{lemma}\label{thm:candidateA}
		For a function $f(\cdot)$ satisfying $	|f(x)-f(x_E)|\leq K|x-x_E|^p$, given the boundary points $x_l$, $x_r$ and their values $f_l\triangleq f(x_l)$ and $f_r\triangleq f(x_r)$, the `candidate' point $x$ between the boundaries $x_l$ and $x_r$ (i.e., $x\in[x_l,x_r]$) with the lowest possible value is the solution of $K|x_r-x|^p-K|x_l-x|^p=f_r-f_l,$ and its score is $s=f_l-K|x_l-x|^p=f_r-K|x_r-x|^p.$\\
		If $x\notin(x_l,x_r)$ (where $x$ may not necessarily have a simpler expression), there exists no candidate.
	\end{lemma}
	\begin{proof}
		The proof is similar to the proof of \cref{thm:candidateH}.
	\end{proof}
	We next study the cumulative regret of \cref{alg} with the application in \cref{thm:candidateA}. 
	\begin{lemma}\label{thm:sampleRegretA}
		
		For a function $f(\cdot)$ satisfying $	|f(x)-f(x_E)|\leq K|x-x_E|^p$, let $x_m$ be the next evaluated candidate point, which is between the boundary points $x_l$, $x_r$ and their corresponding values $f_l\triangleq f(x_l)$ and $f_r\triangleq f(x_r)$. The regret incurred by the evaluation of $x_m$ is
		\begin{align}
			\textstyle f(x_m)-\min_{x\in[0,1]}f(x)\leq& K\bar{x}^p+K(x_r-x_l-\bar{x})^p\\
			&-K(x_r-x_l-2\bar{x})^p,
		\end{align} 
		where $\bar{x}=\min(x_m-x_l,x_r-x_m)$.
	\end{lemma}
	\begin{proof}
		Let $\bar{x}=\min(x_m-x_l,x_r-x_m)$.
		From \cref{thm:candidateA}, we have
		\begin{align}
			s_m\triangleq \min(f_l,f_r)-K\bar{x}^p\leq \min_{x\in[x_l,x_r]}f(x).\label{eq:smK}
		\end{align}
		Similar to the proof of \cref{thm:sampleRegretH}, we have $$s_m\leq \min_{x\in[0,1]}f(x).$$ 
		Following the proof style of \cref{thm:sampleRegretH}, from $	|f(x)-f(x_E)|\leq K|x-x_E|^p$ and \eqref{eq:smK}, we have
		\begin{align}
			f(x_m)&-\min_{x\in[0,1]}f(x)\leq K(\bar{x}^p+(x_r-x_l-\bar{x})^p-(x_r-x_l-2\bar{x})^p),
		\end{align}
		which concludes the proof.
	\end{proof}

	Similar to \cref{thm:rec}, we have a recursive inequality for an upper bound of the cumulative regret.
	\begin{lemma}\label{thm:recA}
		
		We have the following recursive relation for an upper bound on the regret.
		\begin{align}
			R_K([0,1],\bar{T})\leq& K\left(\bar{x}^p+(1-\bar{x})^p-(1-2\bar{x})^p\right)\nonumber\\
			&+x^pR_K([0,1],T_1)\\
			&+(1-x)^pR_K([0,1],T_2),
		\end{align}
		where $\bar{x}=\min(x,1-x)$ for some $x\in[0,1]$ and $T_1+T_2=\bar{T}+1=T-1$.
	\end{lemma}
	\begin{proof}
		Let us observe that the lower bounding algorithm in \cref{alg} is deterministic. Hence, given $K$, function $f(\cdot)$ and interval $[0,1]$ (where $0$ and $1$ are already evaluated), it will always evaluate the same points for known $p$. Let these be $x_1^T$, i.e., $$x_1^T=PS_{HoSm}([0,1],f(\cdot),K,T),$$ for some function $PS_{HoSm}(\cdot)$, which represents the application of \cref{alg} to Holder smooth functions. Let $S_{K}([0,1], x_1^T)$ be the cumulative bound resulting from \cref{thm:sampleRegretA} 
		$$S_{K}([0,1], x_1^T)=K\sum_{t=1}^T\left[\bar{x}_t^p+K(x_{t,r}-x_{t,l}-\bar{x}_t)^p-K(x_{t,r}-x_{t,l}-2\bar{x}_t)^p\right],$$
		where $x_{t,r}=\min_{x:x\in\{x_{-1}^{t-1},x>x_t\}}$, $x_{t,l}=\min_{x:x\in\{x_{-1}^{t-1},x<x_t\}}$, $\bar{x}_t=\min(x_{t,r}-x_t,x_t-x_{t,l})$; and $x_{0}=0$, $x_{-1}=1$ are initially evaluated boundary points of $[0,1]$. Let the maximum bound be $$R_{K}([0,1],T)=\max_{0\leq k\leq K}\max_{f(\cdot):\text{$k$-Holder smooth}}S_{K}([0,1],PS_{HoSm}([0,1],f(\cdot),k,T)).$$ 
		
		The rest of the proof is similar to the proof of \cref{thm:rec} and \cref{thm:recH}.
	\end{proof}
	
	To derive the regret bound, we utilize the following useful result.
	\begin{proposition}\label{thm:p1p2sB}	
		For $x\in(0,0.5]$, $p\geq 1$, we have
		\begin{align}
			&\textstyle\frac{1-(1-2x)^p}{1-(1-x)^p}\leq2,&&
			\textstyle\frac{x^p}{1-(1-x)^p}\leq\frac{2^{-p}}{1-2^{-p}},&&\\
			&\textstyle x^p+(1-x)^p-(1-2x)^{p}\leq 1.
		\end{align}
	\end{proposition}
	\begin{proof}
		Let $x\in(0,\frac{1}{2}]$ and $p\geq 1$. We can show that
		\begin{align}
			2(1-x)^p-(1-2x)^p\leq 1,
		\end{align}
		since the left hand side is decreasing, and attains maximum at $x=0$. Thus,	we have
		\begin{align}
			\frac{1-(1-2x)^p}{1-(1-x)^p}\leq2.
		\end{align}
		Moreover, the derivative of $\frac{x^p}{1-(1-x)^p}$ is nonnegative for $x\in(0,0,5]$, hence, it is nondecreasing. Thus, the maximizer is at boundary $x=\frac{1}{2}$ and we have
		\begin{align}
			\frac{x^p}{1-(1-x)^p}\leq\frac{2^{-p}}{1-2^{-p}}.
		\end{align}
		Furthermore, because of convexity for $p\geq 1$, we have from the boundaries $0$ and $\frac{1}{2}$
		\begin{align}
			x^p+(1-x)^p\leq& \max(1,2^{1-p})\\
			\leq& 1,
		\end{align}
		since $p\geq 1$. This results in		
		\begin{align}
			x^p+(1-x)^p-(1-2x)^{p}\leq 1,
		\end{align}
		and concludes the proof.
	\end{proof}

	Using the recursion in \cref{thm:recA}, we reach the following constant cumulative regret bound.
	\begin{theorem}\label{thm:regretKPS}
		For a $K$-Holder smooth function $f(\cdot)$, the traditional Piyavskii--Shubert variant with its application in \cref{thm:candidateA} has the cumulative regret 
		$\textstyle R_T\leq2K\min\left(\log(4T),({1-\gamma})^{-1}\right),$
		where $p\geq 1$ and $\gamma\triangleq\left({2^p-1}\right)^{-1}\leq 1$.
	\end{theorem}
	\begin{proof}
		We run the algorithm for $T$ evaluations. After evaluating the boundary points $\{0,1\}$, there exist $T-2$ more evaluations. We have 
		$$R_T\leq R_K([0,1],T-2)+K.$$
		Let the next evaluated point after $\{0,1\}$ be $x$. We have from \cref{thm:recA}
		\begin{align}
			R_K([0,1],T-2)\leq& K\left(\bar{x}^p+(1-\bar{x})^p-(1-2\bar{x})^p\right)+x^pR_K([0,1],T_1)+(1-x)^pR_K([0,1],T_2).\nonumber
		\end{align}
		If $x\leq1-x$, we can set $T_1\leq T_2$ to upper bound the regret, hence, $T_1\leq \lfloor(T-2)/2\rfloor$, $T_2\leq T-2$ (we can do the converse if $1-x<x$).
		After adding $K$ to both sides and rearranging, we have
		\begin{align}
			&R_K([0,1],T-2)+K\leq\frac{1-(1-2\bar{x})^p}{1-(1-\bar{x})^p}K +\frac{\bar{x}^p\left[R_K([0,1],\lfloor(T-2)/2\rfloor)+K\right]}{1-(1-\bar{x})^p}.
		\end{align}
		Using \cref{thm:p1p2sB} and the recursive relation, we have
		\begin{align}
			R_K([0,1],T-2)+K
			&\leq2K\min\left(N,\frac{1-\gamma^N}{1-\gamma}\right)+\gamma^{N}\left[R_K([0,1],1)+K\right],
		\end{align}
		for $N=\lceil\log T\rceil$, where $\gamma\triangleq\left(\frac{2^{-p}}{1-2^{-p}}\right)\leq 1$, since $p\geq 1$. Using \cref{thm:sampleRegretA}, \cref{thm:p1p2sB} and adding the initial evaluations $x=0$ and $x=1$, we have
		\begin{align}
			R_T&\leq 2K\min\left(N,\frac{1-\gamma^N}{1-\gamma}\right)+2K\gamma^{N}\\
			&\leq2K\min\left(N+1,\frac{1-\gamma^{N+1}}{1-\gamma}\right)
		\end{align}
		which concludes the proof.
	\end{proof}
	
	\begin{itemize*}
		\item When $p=1$ as in \cref{sec:LcontPS}, we have $\gamma\rightarrow1$. Choosing $K=L$, $N= \log T+1$ gives the regret bound $R_T\leq 2L\log(4T),$ which is the same result as in \cref{thm:regretLPS} for Lipschitz continuous $f(\cdot)$.\\
		\item When $p=2$ as in \cref{sec:LsmoothPS}, we have $\gamma=1/3$. Choosing $N\rightarrow \infty$ and $K=0.5H$ gives, $R_T\leq 1.5H,$ which is looser than \cref{thm:regretHPS} because of the similar proof style to \cref{thm:regretLPS}. \\
		\item We can expand the results to $p\geq 2$. We have $R_T\leq \frac{2^{p+1}-2}{2^p-2}K,$ for an arbitrary $p\geq 2$; and $R_T\leq 3K$ for all $p\geq 2$. 
		For any $p>1$, the cumulative regret is $O(K)$, i.e., independent from $T$.
	\end{itemize*}
	
	\section{Regret analysis of the simpler Piyavskii--Shubert variants}
	In this section, as a design choice, the candidate $x$ in the steps of \cref{alg} is selected differently from the traditional lower-bounding algorithms.
	As opposed to the point with the lowest possible functional value between the boundaries $x_l$ and $x_r$ (which is the intersection of the lines that pass through points $(x_l,f_l)$, $(x_r,f_r)$ with slopes $-L$, $L$ respectively for Lipschitz continuous functions); we directly select it as the middle point of $x_l$ and $x_r$. The exact expression is given in the following.
	\begin{definition}\label{thm:candidateBS}
		For an objective function $f(\cdot)$ satisfying \cref{def:condition}, given the boundaries $x_l$, $x_r$ and their evaluations $f_l$, $f_r$; the candidate $x$ between $x_l$ and $x_r$ (i.e., $x\in(x_l,x_r)$) is given by $x=\frac{1}{2}\left(x_r+x_l\right),$
		which is the middle point of $[x_l,x_r]$ if $|f_r-f_l|\leq d(|x_r-x_l|)$. Otherwise, there does not exist a potential minimizer (from \cref{def:condition}) in $(x_l,x_r)$ and we set no such candidate.
	\end{definition}
	
	Because of the way the potential queries are determined, they can be represented as binary strings which increases memory and communication efficiency. The determination of the candidate $x$ is much easier than the traditional variants, which may require solving additional optimization problems.
	The score $s$ in \cref{alg} is also much easier to determine as opposed to the traditional Piyavskii–Shubert variants. The exact expression of the score $s$ is given in the following.
	
	\begin{lemma}\label{thm:score}
		For an objective $f(\cdot)$ satisfying \cref{def:condition}; given the boundary points $x_l$, $x_r$ and their values $f_l\triangleq f(x_l)$, $f_r\triangleq f(x_r)$, we assign the potential query $x=\frac{1}{2}\left(x_r+x_l\right)$ the following score
		$s=\min(f_l,f_r)-d\left(\frac{x_r-x_l}{2}\right),$ which lower bounds the evaluation of the region $[x_l,x_r]$.	
	\end{lemma}
	\begin{proof}
		Suppose $x$ is a local minimum of $f(\cdot)$ in $[x_l,x_r]$. From \cref{def:condition}, we have
		\begin{align}
			f(x)\geq \max\left(f_l-d(x-x_l),f_r-d(x_r-x)\right).
		\end{align}
		If $$x\leq \frac{x_r+x_l}{2},$$ we have $$f(x)\geq f_l-d(x-x_l)
		\geq f_l-d\left(\frac{x_r-x_l}{2}\right)
		\geq \min(f_l,f_r)-d\left(\frac{x_r-x_l}{2}\right);$$
		else if $$x\geq \frac{x_r+x_l}{2},$$ we have $$f(x)\geq f_r-d(x_r-x)
		\geq f_r-d\left(\frac{x_r-x_l}{2}\right)
		\geq \min(f_l,f_r)-d\left(\frac{x_r-x_l}{2}\right),$$
		which concludes the proof.
	\end{proof}

	In our general regret analysis, we again start by bounding the regret of a single queried point.
	
	\begin{lemma}\label{thm:sampleRegretBS}
		
		For an objective $f(\cdot)$ satisfying \cref{def:condition} with some known $d(\cdot)$, let $x_m$ be the next query; $x_l$, $x_r$ are the boundaries and $f_l\triangleq f(x_l)$, $f_r\triangleq f(x_r)$ are their evaluations. The regret incurred by $x_m$ is bounded as
		$	\textstyle f(x_m)-\min_{x\in[0,1]}f(x)\leq 2d(x_r-x_l).
		$
	\end{lemma}
	\begin{proof}
		Since the score $s_m$ of $x_m$ is a lower bound for the function $f(\cdot)$ between $[x_l,x_r]$, we have $$s_m\triangleq \min(f_l,f_r)-d\left(\frac{x_r-x_l}{2}\right)\leq \min_{x\in[x_l,x_r]}f(x).$$ Because of the way our algorithm works, each candidate is inside an interval possibly containing a local minimum better than the past queries. Since we sample the query with the lowest score, we have
		\begin{align}
			s_m&\triangleq\min(f_l,f_r)-d\left(\frac{x_r-x_l}{2}\right)\nonumber\\
			&\leq \min_{x\in[0,1]}f(x).\label{eq:sm}
		\end{align}
		Without loss of generality let $f_l\leq f_r$. Then, because of \cref{def:condition} and the algorithm, we have $$f(x_m)\leq f_l+d(x'-x_l),$$ for some potential local maximizer $x'$ (where $x_r\geq x'\geq x_m$), which satisfies $$f_l+d(x'-x_l)=f_r+d(x_r-x').$$ Thus, using \eqref{eq:sm}, we have
		\begin{align}
			f(x_m)-\min_{x\in[0,1]}f(x)\leq& d(x'-x_l)+d(x_m-x_l)
			\\\leq& d(x_r-x_l)+d(x_m-x_l)
			\\\leq& 2 d(x_r-x_l),						
		\end{align}
		since $\min(f_l,f_r)=f_l$, $(x_r-x_l)/2=x_m-x_l$, $d(\cdot)$ is nondecreasing and $x'-x_l\leq x_r-x_l$, which concludes the proof.
	\end{proof}

	This result bounds the individual regret of a sampled point $x_m$ with only its boundary values $x_l$ and $x_r$ (irrespective of the functional evaluations $f_l$ and $f_r$); hence, is a worst case bound. 
	Next, we study the cumulative regret of the algorithm with the candidate points and scores in \cref{thm:candidateBS} and \cref{thm:score}, respectively, by deriving the cumulative regret bound up to time $T$.
	\begin{theorem}\label{thm:regretBS}
		For a function satisfying \cref{def:condition}, using the candidate of \cref{thm:candidateBS} and the score of \cref{thm:score} in \cref{alg} results in $R_T\leq d(1)+2\sum_{i=0}^{a}2^{i}d\left(\frac{1}{2^i}\right)$ and $r_T\leq 2d(2(T-1)^{-1}),$
		where $a$ is an integer such that $2^a+B+1=T$ for some integer $1\leq B \leq 2^a$.
	\end{theorem}
	\begin{proof}
		We run the algorithm for $T$ sampling times. After sampling the boundary points $x=0$, $x=1$ and the middle point $x=1/2$, there exists $T-3$ more sampling. For a boundary point pair $(x_l,x_r)$, we sample the middle point
		$$x_m\triangleq\frac{x_l+x_r}{2}$$
		and incur the individual regret
		\begin{align}
			f(x_m)-\min_{x\in[0,1]}f(x)\leq 2d(x_r-x_l).
		\end{align}
		The distance between any new boundary points will be halved. Hence, at the worst case scenario, we will sample the oldest candidate points first. The total regret for $T-3$ sampling will be given by
		\begin{align}
			\tilde{R}_T\leq 2\bigg(2d\left(\frac{1}{2}\right)+4d\left(\frac{1}{4}\right)+\ldots+2^{a-1}d\left(\frac{1}{2^{a-1}}\right)+Bd\left(\frac{1}{2^a}\right)\bigg),\label{eq:RTtilBS}
		\end{align}
		where $1\leq B \leq 2^a$ and $\sum_{i=1}^{a-1}2^i=2^a-2=T-3-B.$
		The regret incurred by the first three samples $x=0$, $x=\frac{1}{2}$ and $x=1$ is given by
		\begin{align}
			f(0)+f\left(\frac{1}{2}\right)+f(1)-3\min_{x\in[0,1]}f(x)&\leq d(x_*)+d\left(\abs{x_*-\frac{1}{2}}\right)+d(1-x_*)
			\\&\leq 3d(1),
		\end{align}
		where $x^*$ is a minimizer of $f(\cdot)$. Hence, the total regret is 
		\begin{align}
			R_T\leq d(1)+2\sum_{i=0}^{a}2^{i}d\left(\frac{1}{2^i}\right).
		\end{align}
		From the worst-case construction in \eqref{eq:RTtilBS}, we have the simple regret
		\begin{align}
			r_T\leq 2d(2(T-1)^{-1}),
		\end{align}
		which concludes the proof.
	\end{proof}
	
	The cumulative regret is strongly related with the condition $d(\cdot)$. Next, we provide various examples. together with the algorithmic implementation and the regret results.
	
	
	\begin{corollary}\label{thm:regretLBS}
		If $f(\cdot)$ is Lipschitz continuous with $L$, i.e., $\abs{f(x)-f(y)}\leq L\abs{x-y},$ and the algorithm is run using the score selection of \cref{thm:score} with the regularity $d(|\cdot|)=L|\cdot|$; it has the regrets $R_T\leq L(2\log T+3)$ and $r_T\leq 4L(T-1)^{-1}$.
	\end{corollary}
	\begin{proof}
		If $f(\cdot)$ is Lipschitz continuous with $L$, i.e., $$\abs{f(x)-f(y)}\leq L\abs{x-y},$$ the subderivative is bounded, i.e., $$-L\leq f'(x)\leq L.$$ Hence, $$|f(x)-f(x_E)|\leq L|x-x_E|,$$
		for any extremum $x_E$. 
		Using \cref{thm:regretBS} together with \cref{def:condition} and $$d(|x-x_E|)=L|x-x_E|,$$ we get
		\begin{align}
			R_T&\leq L+2L\sum_{i=0}^{a}2^i\frac{1}{2^i}
			\\&\leq L+2L(a+1)
			\\&\leq 3L+2La
			\\&\leq 3L+2L\log(T)
			\\&\leq L(2\log T+3),
		\end{align}
		where $a$ is such that $2^a+B+1=T$ for some $1\leq B \leq 2^a$, which concludes the proof.
	\end{proof}

	\begin{corollary}\label{thm:regretHBS}
		If $f(\cdot)$ is differentiable and Lipschitz smooth with $H$, i.e., $\abs{f'(x)-f'(y)}\leq H\abs{x-y},$ and the algorithm is run using the score selection of \cref{thm:score} with the regularity $d(|\cdot|)=0.5H|\cdot|^2$; it has the regrets $R_T \leq 2.5H,$ and $r_T\leq {4}H(T-1)^{-2}$.
	\end{corollary}
	\begin{proof}
		A differentiable function $f(\cdot)$ is Lipschitz smooth with $H$ if $$\abs{f'(x)-f'(y)}\leq H\abs{x-y},$$ thus, the second subderivative is bounded, i.e., $$-H\leq f''(x)\leq H.$$ Hence, we have $$|f(x)-f(x_E)|\leq \frac{1}{2}H|x-x_E|^2,$$
		for any extremum $x_E$ of $f(\cdot)$ since the derivative is zero. 
		Using \cref{thm:regretBS} together with \cref{def:condition} and $$d(|x-x_E|)=\frac{1}{2}H|x-x_E|^2,$$ we get the following:
		\begin{align}
			R_T&\leq \frac{1}{2}H+H\sum_{i=0}^{a}2^{i}\frac{1}{2^{2i}}
			\\&\leq \frac{1}{2}H+H\sum_{i=0}^{a}\frac{1}{2^i}
			\\&\leq \frac{1}{2}H+H\sum_{i=0}^{\infty}\frac{1}{2^i}
			\\&\leq 2.5H,
		\end{align}
		where $a$ is such that $2^a+B+1=T$ for some $1\leq B \leq 2^a$, which concludes the proof.
	\end{proof}

	
	\begin{corollary}\label{thm:regretK}
		If $f(\cdot)$ satisfies $|f(x)-f(x_E)|\leq K|x-x_E|^p,$
		for any local extremum $x_E$ of $f(\cdot)$ with $K>0,p>0$ and the algorithm is run using the score selection of \cref{thm:score} with the regularity $d(|\cdot|)=K|\cdot|^p$; it has regrets $R_T \leq K+2K\frac{2^{(1-p)\log(2T)}-1}{2^{1-p}-1}$ and $r_T\leq 2^{1+p}K(T-1)^{-p}$, for $p\neq 1$.
	\end{corollary}
	\begin{proof}
		Using \cref{thm:regretBS} together with \cref{def:condition} and $$d(|x-x_E|)=K|x-x_E|^p,$$ we get
		\begin{align}
			R_T-K&\leq 2K\sum_{i=0}^{a}2^{i}\frac{1}{2^{pi}}
			\\&\leq 2K\sum_{i=0}^{a}{2^{(1-p)i}}
			\\&\leq 2K\frac{2^{(1-p)(a+1)}-1}{2^{1-p}-1}
			\\&\leq 2K\frac{2^{(1-p)\log(2T)}-1}{2^{1-p}-1},
		\end{align}
		where $a$ such that $2^a+B+1=T$ for some $1\leq B \leq 2^a$, which concludes the proof.
	\end{proof}
	
	We have the following results for the cases investigate in \cref{sec:traditional}:\\
	\begin{itemize*}
		\item When $p=1$, this class of functions also includes the Lipschitz continuous functions in \cref{sec:LcontPS}. For $p\rightarrow1$: we have the result of \cref{thm:regretLBS} with $K=L$. The result follows from substituting the relevant parameters in \cref{thm:regretK} and the application of L'Hospital's rule. \\
		\item When $p=2$, this class of functions also includes the Lipschitz smooth functions given in \cref{sec:LsmoothPS}. For $p=2$: we have the result of \cref{thm:regretHBS} with $K=H/2$. The result follows directly from substituting the relevant parameters in \cref{thm:regretK}. \\
		\item When $1\leq p\leq 2$, this class of functions includes the Holder smooth functions given in \cref{sec:LanalyticPS}. For $p>1$: we have $R_T\leq K\left(1+2(1-2^{1-p})^{-1}\right)$. The result follows with $T\rightarrow\infty$ in \cref{thm:regretK}, since $2^{1-p}<1$. The regret bound is a function of $p$. \\
		\item For all $p> 2$, we have $R_T\leq 5K$. The result follows from bounding $p$ by $2$ with $T\rightarrow\infty$ in \cref{thm:regretK}, since $2^{1-p}<1$. The regret bound does not depend on $p$.
	\end{itemize*}
	
	Moreover, we have the following new result:\\
	\begin{itemize*}
		\item When $0< p\leq 1$, this class of functions includes the Holder continuous functions.
		For $0<p<1$: we have $R_T\leq K\frac{2^{2-p}}{2^{1-p}-1}T^{1-p}$. The result follows by the direct substitution of $2^{(1-p)\log(2T)}$ with $(2T)^{1-p}$ in \cref{thm:regretK}. 
		When $0<p<1$, the regret bound is polynomial in $T$.
	\end{itemize*}
	
	\begin{corollary}\label{thm:regretG}
		If function $f(\cdot)$ satisfies $|f(x)-f(x_E)|\leq g(\abs{x-x_E}),$
		for any local extremum $x_E$ of $f(\cdot)$, where $g(\cdot)$ is nonnegative, convex and $g(0)=0$; and the algorithm is run using the score selection of \cref{thm:score} with the regularity $d(|\cdot|)=g(|\cdot|)$; it has the regrets $R_T\leq g(1)\left(2\log(T)+3\right),$ and $r_T\leq 2g(2(T-1)^{-1})$.
	\end{corollary}
	\begin{proof}
		Using \cref{thm:regretBS} and \cref{def:condition} with convex regularity, we get the following:
		\begin{align}
			R_T-g(1)&\leq 2\sum_{i=0}^{a}2^{i}g\left(\frac{1}{2^i}\right)
			\\&\leq 2\sum_{i=0}^{a}g\left(\frac{2^i}{2^i}\right)
			\\&\leq 2(a+1)g(1)
			\\&\leq 2g(1)\log(T)+2g(1)
		\end{align}
		since $g(\cdot)$ is convex and $g(0)=0$,
		where $a$ such that $2^a+B+1=T$ for some  $1\leq B \leq 2^a$, which concludes the proof.
	\end{proof}
	
	
	If we have different convex regularities in unknown regions of the domain, we can model the regularity as the maximum of these regularities and achieve at most logarithmic regret since the maximum of convex functions is convex. Moreover, since \cref{thm:regretG} holds for any convex regularity, we have $R_T \leq K\left(1+2\min\left(\log(2T),(1-2^{1-p})^{-1}\right)\right),$ for any $T$, $p\geq 1$ after combining with \cref{thm:regretK}.
	\begin{corollary}\label{thm:regretC}
		If function $f(\cdot)$ satisfies $|f(x)-f(x_E)|\leq h(x-x_E),$
		for any local extremum $x_E$ of $f(\cdot)$, where $h(\cdot)$ is nonnegative, concave and $h(0)=0$; and the algorithm is run using the score selection of \cref{thm:score} with the regularity $d(|\cdot|)=h(|\cdot|)$; it has the regrets $R_T\leq4Th\left(\frac{\log(4T)}{2T}\right),$ and $r_T\leq2h(2(T-1)^{-1})$.
	\end{corollary}
	\begin{proof}
		Using \cref{thm:regretBS} and \cref{def:condition} with concave regularity, we get	
		\begin{align}
			R_T&\leq h(1)+  \sum_{i=0}^{a}2^{i+1}h\left(\frac{1}{2^i}\right)
			\\&\leq(2^{a+2}-1)h\left(\frac{1}{2^{a+2}-1}+\sum_{i=0}^{a}\frac{2}{2^{a+2}-1}\right)
			\\&\leq(2^{a+2}-1)h\left(\frac{2a+3}{2^{a+2}-1}\right)
			\\&\leq(2^{a+2}+4B+4)h\left(\frac{2a+4}{2^{a+2}+4B+4}\right)
			\\&\leq4Th\left(\frac{2a+4}{4T}\right)
			\\&\leq4Th\left(\frac{\log(4T)}{2T}\right)
		\end{align}
		from Jensen's inequality and concavity of $h(\cdot)$, where $a$ such that $2^a+B+1=T$ for some  $1\leq B \leq 2^a$.
	\end{proof}

	\begin{theorem}[]\label{thm:minimax}
		For convex or concave regularities, our regret optimality gap is at most logarithmic.
	\end{theorem}
	\begin{proof}
		For a concave regularity $h(\cdot)$, suppose $f(\cdot)$ is such that, it has a minimum $f^*$ at some $x^*$ and its value increases in both sides with the concave function $h(\cdot)$ around some $\epsilon$-neighborhood. Hence,
		\begin{align}
			f(x)=\begin{cases}
				f^*+h(\epsilon),& x\in[x^*+\epsilon,1]\\
				f^*+h(x-x^*),& x\in[x^*,x^*+\epsilon)\\
				f^*+h(x^*-x),& x\in[x^*-\epsilon,x^*)\\
				f^*+h(\epsilon),& x\in[0,x^*-\epsilon)\\
			\end{cases}.\label{eq:mmf}
		\end{align}
		We point out that the function $f(\cdot)$ in \eqref{eq:mmf} satisfies the regularity in \cref{def:condition}. Let the evaluation points of an arbitrary algorithm be $\{x_t\}_{t=1}^T$. Let the $\epsilon$ be such that the algorithm evaluations are always the same and equal to $$f(x_t)=f^*+h(\epsilon)$$ for all $t$ and some $\epsilon$, which will make the evaluations uninformative for any algorithm. The cumulative regret for this particular case will be given by $$R_{T}^{mm}\triangleq \sum_{t=1}^T(f(x_t)-f^*)=Th(\epsilon).$$ To maximize the regret, the adversary maximizes $\epsilon$ by selecting the most suitable $x^*$. On the other hand, to minimize the regret, the algorithm minimizes $\epsilon$ by selecting the most suitable $\{x_t\}_{t=1}^T$. Hence, the minimax $\epsilon$ is given by $$\epsilon^*=\min_{\{x_t\}_{t=1}^T}\max_{x^*\in[0,1]}\min_t\abs{x^*-x_t}.$$ Since the minimization $\{x_t\}_{t=1}^T$ equally splits $[0,1]$, we have $$\epsilon^*=\frac{1}{2T}.$$ Thus, the minimax regret lower bound is given by 
		\begin{align}
			R_T^{mm}=& Th\left(\frac{1}{2T}\right),\\\geq& \frac{1}{2}Th\left(T^{-1}\right).
		\end{align}
		Since for concave regularities, we achieve $O(T\log(T)h(T^{-1}))$ cumulative regret (from \cref{thm:regretC}), our algorithm has at most a logarithmic optimality gap. With the same example, the simple regret will be lower bounded by $h\left(\frac{1}{2T}\right)$. Hence, our simple regret is minimax optimal.
		
		For a convex regularity $g(\cdot)$, we achieve at most $O(\log T)$ cumulative regret from \cref{thm:regretG}. Since even a single evaluation incurs $\Omega(1)$ regret, our minimax optimality gap is at most logarithmic. With a similar analysis to the concave regularity, our simple regret is minimax optimal if there exists a finite constant $c_2$ such that $g(2x)\leq c_2g(x)$, which is obviously true for polynomial regularities. 
	\end{proof}
	
	\section{Extensions}
	\subsection{Implementation on multivariate scenarios}\label{sec:multi}
	We can straightforwardly extend the algorithm for use in multivariate global optimization when the regularity condition is on the function itself and not just its extrema. As an example, we will investigate the Holder continuous functions, which satisfy the following condition: $|f(x)-f(y)|\leq \norm{x-y}_\infty^\alpha,$
	for any $x,y\in[0,1]^d$ (where $d$ is the dimension) and $\alpha>0$. 
	To cast the multivariate optimization problem of $\min_{x\in[0,1]^d}f(x)$ to a univariate optimization problem, we can fill the optimization region $[0,1]^d$ with hyper-cubes of dimension $d$ and edge length $\epsilon$. For suitable $\epsilon$, the total number of these hyper-cubes are $N=O(\epsilon^{-d})$. Then, we connect the centers of these hyper-cubes with a non-overlapping path. The total length of this curve is $D\leq O(\epsilon^{1-d})$. We observe that 
	if we treat this path as our univariate optimization domain, the Holder condition still holds because of the triangle inequality.
	
	\begin{theorem}
		When optimizing a function $f(\cdot)$ over $[0,1]^d$ with the regularity $|f(x)-f(y)|\leq \norm{x-y}_\infty^\alpha, \forall x,y\in[0,1]^d$, we have the cumulative regret
		${R}_T\leq O(T^{1-\frac{\alpha}{d}}),
		$
		for time horizon $T$, when the centers of hyper-cubes with length $\epsilon=O(T^{-\frac{1}{d}})$ are connected with a non-overlapping path.
	\end{theorem}
	\begin{proof}
		For a time horizon $T$, diameter $D$ (length of the non-overlapping path over the hyper-cube centers) and condition $|f(x)-f(y)|\leq \norm{x-y}_\infty^\alpha, \forall x,y,\in[0,1]^d$, we have
		\begin{align}
			\tilde{R}_T=O(D^\alpha T^{1-\alpha}),
		\end{align} 
		regret with respect to any point on the curve from \cref{thm:regretK} since the connection of these hyper-cube centers constructs a one-dimensional line. 
		
		The diameter $D$ here is determined by the choice of $\epsilon$, i.e., the parameter that decides the side length of an hyper-cube, as follows. First, fix some $\epsilon$. Then, split the set $[0,1]^d$ into $(1/\epsilon')^d$ hyper-cubes with lengths $\epsilon'$, where $\epsilon' = 1 / \lfloor 1/\epsilon \rfloor$. Here, $\epsilon'$ is such that $\epsilon \leq \epsilon' < 2\epsilon$, since $(1/\epsilon)\geq \lfloor 1/\epsilon \rfloor\geq (0.5/\epsilon)$ for $\epsilon\leq 1$. 
		
		The optimal point is at most $\epsilon'/2$ away from the curve (i.e., its distance to the line, which pierces the hyper-cube that contains it, is less than $\epsilon$), we have
		\begin{align}
			{R}_T\leq& O(D^\alpha T^{1-\alpha}+T\epsilon^\alpha)\\
			\leq&O(\epsilon^{\alpha(1-d)}T^{1-\alpha}+T\epsilon^\alpha),
		\end{align} 
		where the first inequality represent the regret components from firstly on the line (from the hyper-cube path) and secondly off the line (accumulated per turn distances to the insides of respective hyper-cubes), and the second inequality is achieved by replacing $D^\alpha = \left(\epsilon' (1/\epsilon')^d\right)^\alpha = (\epsilon')^{\alpha(1-d)} \leq \epsilon^{\alpha(1-d)}$ since $\epsilon\leq\epsilon', d\geq 1, \alpha\geq 0, T\geq 0$. 
		Then, if we choose $\epsilon=T^{-\frac{1}{d}}$, we have the regret
		\begin{align}
			{R}_T\leq&O(T^{1-\frac{\alpha}{d}}),
		\end{align}
		which is minimax optimal using similar arguments in \cref{thm:minimax}).
		
		\textbf{Example} to the non-overlapping path that connect these $(1/\epsilon')^d$ hyper-cubes can be as follows. Denote the position of each hyper-cube with the integers $[i_1, \ldots, i_d]$, where $i_k \in \{1, \ldots, K\}$ for $k \in \{1,\ldots,d\}$ and $K = (1/\epsilon') = \lfloor 1/\epsilon \rfloor$. Notice that any pair of adjacent (with a common facet) hyper-cubes have the same indices $i_k$ at all but one $k$, and the differing index only differs by $1$. Hence, the hyper-cubes can be spanned similar to the following construct. 
		\begin{itemize}
			\item Initialize the increment indicator $a_1 = [1, \ldots, 1]$ and the starting hyper-cube $c_1 = [1,\ldots,1]$. 
			
			\item For $n = 1, \ldots, K^d - 1$:
			\begin{itemize}
				\item Fill as $c_{n+1} = c_n$ and $a_{n+1}=a_n$ temporarily.
				\item For $k = 1, \ldots, d$:
				\begin{itemize}
					\item If (($a_{n,k} = 1$ and $c_{n,k} < K$) or ($a_{n,k} = -1$ and $c_{n,k} > 1$)):
					\\$c_{n+1,k} = c_{n,k} + a_{n,k}$.
					\\break loop.
					\item Else: 
					\\$a_{n+1,k} = -a_{n,k}$.
				\end{itemize}		
			\end{itemize}  
		\end{itemize}
	\end{proof}
	
	\subsection{Dealing with noisy observations}
	Suppose each evaluation $f_t=f(x_t)$ has an additive noise $v_t$, i.e., we observe $\tilde{f}_t=f_t+v_t$. Because of the way the algorithm works, this noise will result in an additive component in the sample regret at every time $t\geq 3$, since the boundary points are initially sampled without comparisons. Since the scores of our candidate points are compared against each other, which depend on the evaluations, the redundant regret at time $t\geq 3$ will be at most $\max_{\tau<t}v_\tau-\min_{\tau<t}v_\tau.$ In the cumulative regret bound, we will have an additive component $\sum_{t=3}^T\left(\max_{\tau<t}v_\tau-\min_{\tau<t}v_\tau\right).$ Let us assume $v_t$ are independent identically distributed Gaussian noise with an unknown mean $\mu$ and unknown variance $\sigma^2$. The expectation of the range of normal random variables over $t$ samples is bounded by $O(\sigma\sqrt{\log t})$. Hence, the summation will be $O(\sigma T\sqrt{\log T})$. Obviously, this is not sublinear. We can circumvent this problem by evaluating each candidate $K$ times and utilize the average of samples as its evaluation. 
	\begin{theorem}
		Given the noise-free regret $\tilde{O}(T^{1-\alpha})$ for $0\leq\alpha\leq 1$ and i.i.d. Gaussian sample noise with unknown mean $\mu$ and unknown variance $\sigma^2$; the expected cumulative regret is
		$\E[R_T]\leq\tilde{O}\left(T^{\frac{1+\alpha}{1+2\alpha}}\right),$ when each query is sampled $T^{\frac{2\alpha}{1+2\alpha}}$ times and their average is utilized as the evaluation. 
	\end{theorem}
	\begin{proof}
		Let the original noise-free regret bound be $\tilde{O}(T^{1-\alpha})$ with $0\leq\alpha\leq 1$. 
		
		Each noisy evaluation $\tilde{f}_t=f_t+v_t$, has an i.i.d. Gaussian sample noise $v_t$ with unknown mean $\mu$ and unknown variance $\sigma^2$. Since we sample each query $K$ times, the average noise $\tilde{v}_t$ will have a variance $\sigma^2/K$. The expectation of the range of normal random variables over $t$ samples is bounded by $O(\sigma\sqrt{\log t})$. Hence, in the regret bound, we will have an additional expected redundancy 
		\begin{align}
			\E\left[\max_{\tau<t}\tilde{v}_\tau-\min_{\tau<t}\tilde{v}_\tau\right]\leq O\left(\frac{\sigma}{\sqrt{K}}\log t\right).
		\end{align}
		for each distinct query. In a time horizon $T$, since each query is sampled $K$ times, we evaluate $T/K$ number of queries. Hence, the cumulative expected noise redundancy will be
		\begin{align}
			O\left(\frac{\sigma}{\sqrt{K}}\frac{T}{K}\log\left(\frac{T}{K}\right)\right).
		\end{align}
		Moreover, the regret of $T/K$ queries are given by $(T/K)^{1-\alpha}$. Since each query is sampled $K$ times, we have the following expected regret for a time horizon $T$.
		\begin{align}
			\E[R_T]\leq& {O}\left(K\left(\frac{T}{K}\right)^{1-\alpha}\right)+\tilde{O}\left(K\frac{\sigma}{\sqrt{K}}\frac{T}{K}\right)
			\\\leq&\tilde{O}\left(K^\alpha T^{1-\alpha}+K^{-\frac{1}{2}}T\right)
		\end{align}
		since $\sigma$ is constant. Optimizing with $K=T^{\frac{2\alpha}{1+2\alpha}}$, we have
		\begin{align}
			\E[R_T]\leq\tilde{O}(T^{\frac{1+\alpha}{1+2\alpha}})
		\end{align}
		which is sublinear for $\alpha>0$ and concludes the proof.

	\end{proof}
	
	Similar derivations can be made for different classes of noises.
	\subsection{Regret for unknown regularity}
	When the regularity is unknown, we run the algorithm as in \cref{thm:regretLBS}. 
	\begin{definition}\label{def:sel}
		Given the boundaries $x_l$, $x_r$ and their evaluations $f_l\triangleq f(x_l)$, $f_r\triangleq f(x_r)$; the candidate $x=\frac{1}{2}\left(x_r+x_l\right)$ has the score
		$s=\min(f_l,f_r)-L_0\left|({x_r-x_l})/{2}\right|,$ for and input $L_0>0$.
	\end{definition}
	
	We point out that because of the working structure of the algorithm, if all the scores were offset by some $\epsilon$, the sampled points would not change since the candidate scores are compared against each other. Hence, the algorithm would sample the same candidate points with the same evaluations if the score were set as $s=\min(f_l,f_r)-L_0\left|({x_r-x_l})/{2}\right|-\epsilon,$ for any $\epsilon$. 
	In our general regret analysis, we start by bounding the regret of a single queried point.
	\begin{proposition}\label{thm:sampleRegretU}
		
		For an objective $f(\cdot)$ satisfying \cref{def:condition} with some unknown $d(\cdot)$, and input $L_0$; let $x_m$ be the next sampled query, which is the middle of the boundary points $x_l$, $x_r$ together with their corresponding evaluations $f_l\triangleq f(x_l)$ and $f_r\triangleq f(x_r)$. The regret incurred by $x_m$ is
		\begin{align}
			\textstyle f(x_m)-\min_{x\in[0,1]}f(x)\leq d(x_r-x_l)+L_0\left(\frac{x_r-x_l}{2}\right)+\epsilon_0,	
		\end{align}
		where $
		\epsilon_0=\max_{0\leq\triangle\leq 1}\left(d(\triangle)-L_0\triangle\right).$
	\end{proposition}
	\begin{proof}
		Let $\epsilon_0=\max_{0\leq\triangle\leq 1}\left(d(\triangle)-L_0\triangle\right).$ Since $\epsilon_0$ is the maximum discrepancy between the true redundancy and our input redundancy $L_0|\cdot|$, the shifted score $s_m-\epsilon_0$ of $x_m$ is a lower bound for the function $f(\cdot)$ between $[x_l,x_r]$, we have $$s_m-\epsilon_0\triangleq \min(f_l,f_r)-L_0\left(\frac{x_r-x_l}{2}\right)-\epsilon_0\leq \min_{x\in[x_l,x_r]}f(x).$$ Because of the way our algorithm works, each candidate is inside an interval possibly containing a local minimum better than the past queries. Since we sample the query with the lowest score, we have
		\begin{align}
			s_m\triangleq\min(f_l,f_r)-L_0\left(\frac{x_r-x_l}{2}\right)\leq \min_{x\in[0,1]}f(x)+\epsilon_0.\label{eq:smE}
		\end{align}
		Without loss of generality let $f_l\leq f_r$. Then, because of \cref{def:condition} and the algorithm, we have $$f(x_m)\leq f_l+d(x'-x_l),$$ for some potential local maximizer $x'$ (where $x_r\geq x'\geq x_m$), which satisfies $$f_l+d(x'-x_l)=f_r+d(x_r-x').$$ Thus, using \eqref{eq:smE}, we have
		\begin{align}
			f(x_m)-\min_{x\in[0,1]}f(x)\leq& d(x'-x_l)+L_0\left(\frac{x_r-x_l}{2}\right)+\epsilon_0
			\\\leq& d(x_r-x_l)+L_0\left(\frac{x_r-x_l}{2}\right)+\epsilon_0,					
		\end{align}
		since $\min(f_l,f_r)=f_l$, $d(\cdot)$ is nondecreasing and $x'-x_l\leq x_r-x_l$, which concludes the proof.
	\end{proof}
	
	
	Next, we study the cumulative regret of the algorithm with the candidate points and scores in \cref{thm:candidate} and \cref{thm:score}, respectively, by deriving the cumulative regret bound up to time $T$.
	
	\begin{theorem}\label{thm:regretU}
		For a function $f(\cdot)$ satisfying \cref{def:condition} running \cref{alg} with the selections in \cref{def:sel} results in $R_T\leq2d(1)+\sum_{i=0}^{a}2^{i}d\left(\frac{1}{2^i}\right)+L_0 a+\epsilon_0T,r_T\leq d(2(T-1)^{-1})+2L_0(T-1)^{-1}+\epsilon_0,$
		where $a$ is an integer such that $2^a+B+1=T$ for some integer $1\leq B \leq 2^a$.
	\end{theorem}
	\begin{proof}
		We run the algorithm for $T$ sampling times. After sampling the boundary points $x=0$, $x=1$ and the middle point $x=1/2$, there exists $T-3$ more sampling. For a boundary point pair $(x_l,x_r)$, we sample the middle point
		$x_m\triangleq\frac{x_l+x_r}{2}$
		and incur the individual regret
		\begin{align}
			f(x_m)-\min_{x\in[0,1]}f(x)\leq d(x_r-x_l)+L_0\left(\frac{x_r-x_l}{2}\right)+\epsilon_0.	
		\end{align}
		The distance between any new boundary points will be halved. Hence, at the worst case scenario, we will sample the oldest candidate points first. The total regret for $T-3$ sampling will be given by
		\begin{align}
			\tilde{R}_T\leq& \bigg(2d\left(\frac{1}{2}\right)+4d\left(\frac{1}{4}\right)+\ldots+2^{a-1}d\left(\frac{1}{2^{a-1}}\right)+Bd\left(\frac{1}{2^a}\right)\bigg),\\&+L_0\bigg(2\left(\frac{1}{2}\right)+4\left(\frac{1}{4}\right)+\ldots+2^{a-1}\left(\frac{1}{2^{a-1}}\right)+B\left(\frac{1}{2^a}\right)\bigg)
			\\&+\epsilon_0(T-3)			
			\label{eq:RTtil}
		\end{align}
		where $1\leq B \leq 2^a$ and $\sum_{i=1}^{a-1}2^i=2^a-2=T-3-B.$
		The regret incurred by the first three samples $x=0$, $x=\frac{1}{2}$ and $x=1$ is given by
		\begin{align}
			f(0)+f\left(\frac{1}{2}\right)+f(1)-3\min_{x\in[0,1]}f(x)&\leq d(x_*)+d\left(\abs{x_*-\frac{1}{2}}\right)+d(1-x_*)
			\\&\leq 3d(1),
		\end{align}
		where $x^*$ is a minimizer of $f(\cdot)$. Hence, the total regret is 
		\begin{align}
			R_T\leq& 2d(1)+\sum_{i=0}^{a}2^{i}d\left(\frac{1}{2^i}\right)+L_0\sum_{i=1}^a2^i\left(\frac{1}{2^i}\right)+\epsilon_0(T-3),
			\\\leq&2d(1)+\sum_{i=0}^{a}2^{i}d\left(\frac{1}{2^i}\right)+L_0 a+\epsilon_0T,
		\end{align}
		From \eqref{eq:RTtil}, we have
		\begin{align}
			r_T\leq d(2(T-1)^{-1})+2L_0(T-1)^{-1}+\epsilon_0,
		\end{align}
		which concludes the proof.
	\end{proof}

	This result differs from \cref{thm:regretBS} by its dependence of $L_0$ and $\epsilon_0$.
	
	
	\begin{theorem}
		If $d(\cdot)$ is convex, we have regret $R_T\leq(3+\log T)d(1)+L_0\log T+(d(1)-L_0)^+ T$, which is piecewise linear with $L_0$.
	\end{theorem}
	\begin{proof}
		From \cref{thm:regretU}, the algorithm has the following regrets $$R_T\leq2d(1)+\sum_{i=0}^{a}2^{i}d\left(\frac{1}{2^i}\right)+L_0 a+\epsilon_0T,$$
		where $a$ is an integer such that $2^a+B+1=T$ for some integer $1\leq B \leq 2^a$. 
		If $d(\cdot)$ is convex, we have $\epsilon_0=(d(1)-L_0)^+.$
		Hence,
		\begin{align}
			R_T\leq& 2d(1)+(a+1)d(1)+L_0a+\epsilon_0T,
			\\\leq&(3+\log T)d(1)+L_0\log T+\epsilon_0 T
			\\\leq&(3+\log T)d(1)+L_0\log T+(d(1)-L_0)^+ T
		\end{align}
		Hence, $R_T$ is a function of the input $L_0\geq 0$ such that $$R_T\leq (3+\log T)d(1)+F(L_0),$$ where $F(\cdot)$ is piecewise linear such that $F(\cdot)$ decreases from $0$ to $d(1)$ and increases from $d(1)$ onwards. It has the mappings: $F(0)=Td(1)$, $F(d(1))=d(1)\log T$, $F(L_0)=L_0\log T$ for $L_0> d(1)$.
	\end{proof}
	
	
	\begin{theorem}\label{thm:concave}
		If $d(\cdot)$ is concave, we have $R_T\leq2Td\left(\frac{1}{2T}\right)+L_0\log T+\epsilon_0T,$ 
		\\where
		$\epsilon_0= \begin{cases}
			d(\triangle_0)-L_0\triangle_0,& \text{$\exists \triangle_0: 0\leq\triangle_0\leq1$ and $d'(\triangle_0)=L_0$}\\
			(d(1)-L_0)^+,& \text{otherwise}
		\end{cases}.$
	\end{theorem}
	\begin{proof}
		From \cref{thm:regretU}, the algorithm has the following regrets $$R_T\leq2d(1)+\sum_{i=0}^{a}2^{i}d\left(\frac{1}{2^i}\right)+L_0 a+\epsilon_0T,$$
		where $a$ is an integer such that $2^a+B+1=T$ for some integer $1\leq B \leq 2^a$. If $d(\cdot)$ is concave, we have
		\begin{align}
			R_T\leq& (2^{a+1}+1)d\left(\frac{1}{2^{a+1}+1}\right)+L_0a+\epsilon_0T,
			\\\leq& (2^{a+1}+2B+2)d\left(\frac{1}{2^{a+1}+2B+2}\right)+L_0a+\epsilon_0T,
			\\\leq& 2Td\left(\frac{1}{2T}\right)+L_0\log T+\epsilon_0T.
		\end{align}
		Since $
		\epsilon_0=\max_{0\leq\triangle\leq 1}\left(d(\triangle)-L_0\triangle\right),$
		when $d(\cdot)$ is concave, we have
		$$\epsilon_0= \begin{cases}
			(d(1)-L_0),&L_0\leq\min_{0\leq\triangle\leq 1}d'(\triangle)=d'(1)\\
			0,& L_0\geq \max_{0\leq\triangle\leq 1}d'(\triangle)=d'(0)\\
			d(\triangle_0)-L_0\triangle_0,& \text{otherwise, for some $\triangle_0$, $0\leq\triangle_0\leq1$ such that $d'(\triangle_0)=L_0$}\\
		\end{cases},$$
		which concludes the proof.
	\end{proof}
	
	If there exist no $0\leq\triangle_0\leq1$ such that $d'(\triangle_0)=L_0$; the regret dependency on $L_0$ will be similar as in the convex regularity, where the regret $R_T$ will be the base concave regret plus a piecewise linear term with the same behavior. More interesting case is the first one, where there exists $0\leq\triangle_0\leq1$ such that $d'(\triangle_0)=L_0$. For presentation, let us assume $d(\cdot)=C|\cdot|^p$ for some $C>0$ and $1> p> 0$.
	
	\begin{corollary}
		If $d(\cdot)=C|\cdot|^p$ and $p<1$, we have $R_T\leq C(2T)^{1-p}+T^{1-p_0}\log T+T^{1-\frac{(1-p_0)p}{1-p}}C^{\frac{1}{1-p}}$ for $p_0=1-\log(L_0)/\log(T)$. The logarithmic regret $\log(R_T)$ is piecewise linear with $\log(L_0)$.
	\end{corollary}
	\begin{proof}
		From $d'(\triangle_0)=L_0$, we have 
		\begin{align}
			L_0=&Cp\triangle_0^{p-1},\\
			\triangle_0=&\left(\frac{L_0}{Cp}\right)^{\frac{1}{p-1}}.
		\end{align}
		Thus,
		\begin{align}
			\epsilon_0=&d(\triangle_0)-L_0\triangle_0,
			\\=&C\left(\frac{L_0}{Cp}\right)^{\frac{p}{p-1}}-L_0\left(\frac{L_0}{Cp}\right)^{\frac{1}{p-1}},\\
			=&L_0^{\frac{p}{p-1}}C^{\frac{1}{1-p}}\left(p^{\frac{p}{1-p}}-p^{\frac{1}{1-p}}\right),\\
			\leq&L_0^{\frac{p}{p-1}}C^{\frac{1}{1-p}}
		\end{align}
		since $p<1$. From \cref{thm:concave},
		\begin{align}
			R_T\leq C(2T)^{1-p}+L_0\log T+L_0^{-\frac{p}{1-p}}C^{\frac{1}{1-p}}T,
		\end{align}
		
		Let $L_0=T^{1-p_0}$ for some $p_0$. We have 
		\begin{align}
			R_T\leq& C(2T)^{1-p}+T^{1-p_0}\log T+T^{-(1-p_0)\frac{p}{1-p}}C^{\frac{1}{1-p}}T,\\
			\leq&\tilde{O}\left(T^{G(p_0)}\right)
		\end{align}
		for some piecewise linear function $G(p_0)$, which is $G(0)=G(1)=1$ and minimum at $G(p)$, i.e.,
		\begin{align}
			G(p_0)=\begin{cases}
				{1-{p_0}},&0<p_0\leq p,
				\\{\frac{1-2p}{1-p}+p_0\frac{p}{1-p}},&p<p_0<1,
			\end{cases},
		\end{align}
		which concludes the proof.
	\end{proof}

	\section{Limitations}
	The limitations in our approach are: (1) the regret results are minimax optimal when the regularity condition is known, (2) for noisy evaluations, the noise is assumed to be additive and independent, (3) for minimax optimal regret in multivariate scenarios, the regularity is on the whole function in addition to its extrema, (4) for unknown regularities, sublinear regret is achieved when the regularity prediction is sufficiently close to the true regularity. 
	
	\section{Conclusion}\label{sec:conclusion}
	We studied the problem of univariate global optimization and investigated the cumulative regret of the Piyavskii--Shubert algorithm and its variants. Instead of the traditional Lipschitz regularity, we consider the weaker extrema-specific regularity condition, which allows for a much larger class of functions in addition to the Lipschitz continuous, Lipschitz smooth and Holder continuous functions. We showed that the Piyavskii--Shubert variants have minimax optimal cumulative regrets.
	We also showed that simpler Piyavskii--Shubert variants with predetermined queries perform just as well as the traditional variants. We extended our analyses to general convex and concave regularities. Finally, we considered further extensions by investigating the performance of the Piyavskii-Shubert variants in the scenarios with unknown regularity, noisy evaluation and multivariate domain.

	
	\scriptsize
	\bibliographystyle{ieeetran}
	\bibliography{double_bib}	
	
	\normalsize
	
\end{document}